\pdfoutput=1

\documentclass[11pt]{article}

\usepackage{coling}

\usepackage{times}
\usepackage{latexsym}
\usepackage{booktabs} 
\usepackage[T1]{fontenc}

\usepackage[utf8]{inputenc}

\usepackage{microtype}
\usepackage{colortbl}
\usepackage{xcolor}
\usepackage{inconsolata}
\usepackage{newfloat}
\usepackage{algorithmic}
\usepackage{algorithm}
\usepackage{listings}
\usepackage{times}  
\usepackage{helvet}  
\usepackage{courier}  
\usepackage{graphicx} 
\usepackage{natbib}  
\usepackage{caption} 
\usepackage{algorithm}
\usepackage{algorithmic}

\usepackage{amsmath}
\usepackage{amsfonts}
\usepackage{amssymb}
\usepackage{amsthm, amsmath}
\usepackage{mathrsfs}
\usepackage{indentfirst}
\usepackage{multirow}
\usepackage{graphicx} 
\usepackage{color} 
\usepackage{subcaption}

\title{A Compressive Memory-based Retrieval Approach for Event Argument Extraction}

\author{Wanlong Liu\textsuperscript{\rm 1},
    Enqi Zhang\textsuperscript{\rm 1},
    \textbf{Li Zhou}\textsuperscript{\rm 1}, 
    Dingyi Zeng\textsuperscript{\rm 1},
    \textbf{Shaohuan Cheng}\textsuperscript{\rm 1}, \\
 \textbf{Chen Zhang}\textsuperscript{\rm 2},
    \textbf{Malu Zhang}\textsuperscript{\rm 1}, 
    \textbf{Wenyu Chen}\textsuperscript{\rm 1}
    \\
        {\textsuperscript{\rm 1}{University of Electronic Science and Technology of China}} \\ {\textsuperscript{\rm 2}{	National University of Singapore}}\\liuwanlong@std.uestc.edu.cn, maluzhang@uestc.edu.cn,
        cwy@uestc.edu.cn}

\begin{document}
\maketitle
\begin{abstract}
Recent works have demonstrated the effectiveness of retrieval augmentation in the Event Argument Extraction (EAE) task. However, existing retrieval-based EAE methods have two main limitations: (1) input length constraints and (2) the gap between the retriever and the inference model. These issues limit the diversity and quality of the retrieved information.
In this paper, we propose a \textbf{C}ompressive \textbf{M}emory-based \textbf{R}etrieval (CMR) mechanism for EAE, which addresses the two limitations mentioned above. Our compressive memory, designed as a dynamic matrix that effectively caches retrieved information and supports continuous updates, overcomes the limitations of the input length. Additionally, after pre-loading all candidate demonstrations into the compressive memory, the model further retrieves and filters relevant information from memory based on the input query, bridging the gap between the retriever and the inference model.
Extensive experiments show that our method achieves new state-of-the-art performance on three public datasets (RAMS, WikiEvents, ACE05), significantly outperforming existing retrieval-based EAE methods.

\end{abstract}

\section{Introduction}
\begin{figure}[t]
    \centering
    \includegraphics[width=\linewidth]{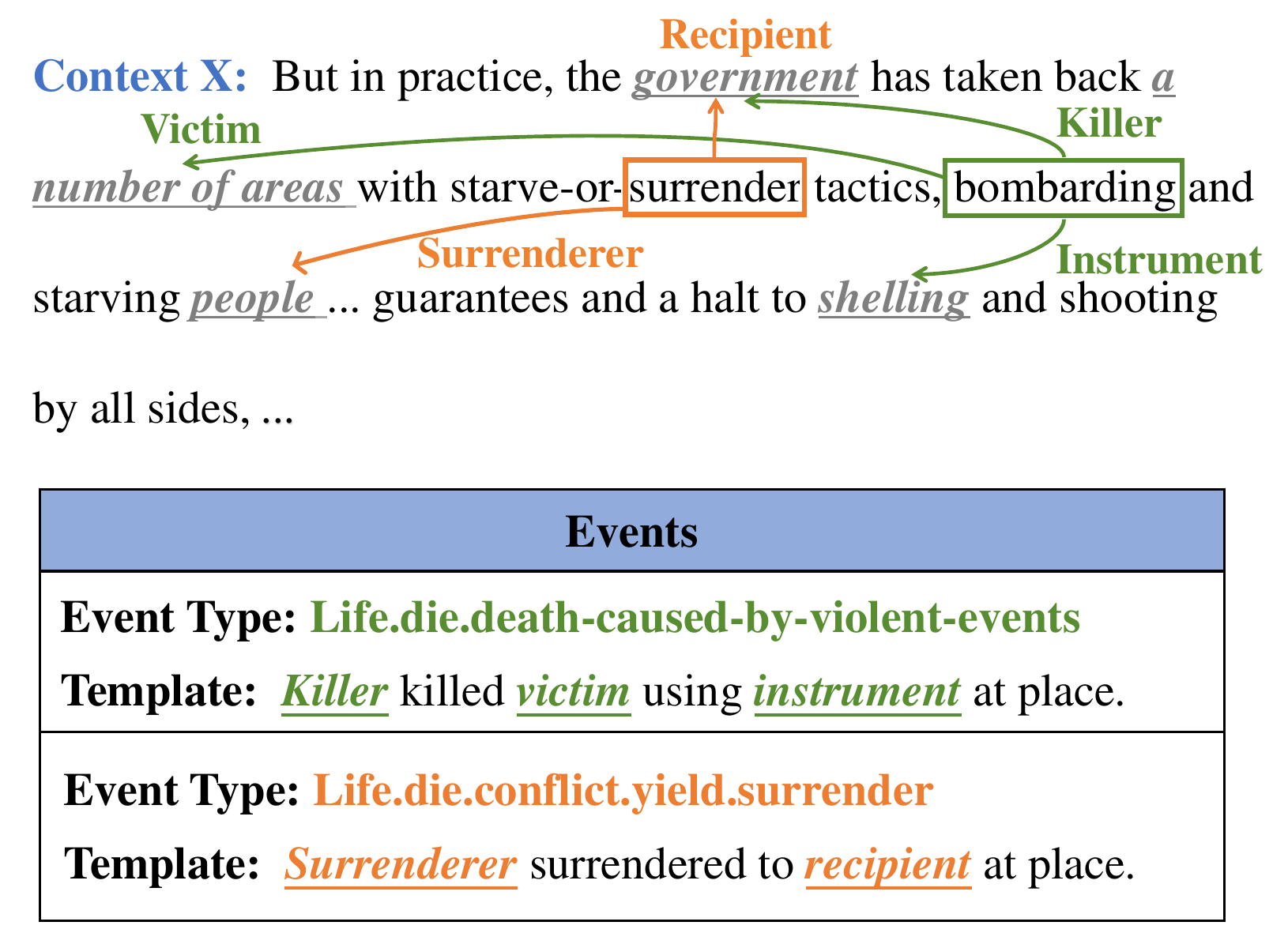}
    \caption{An example of an EAE task from the RAMS dataset~\cite{ebner2020multi}. Each underlined section in the template (prompt), known as a role slot, corresponds to a specific argument role.}
    \label{fig1}
\end{figure}

Event argument extraction (EAE) is a crucial and challenging subtask of event extraction~\cite{ren2022clio, yang2021document}, aimed at identifying event-related arguments and determining their roles within texts. 
For instance, as shown in Figure~\ref{fig1}, when the target event is \texttt{Life.die.death} with the trigger \textit{bombarding}, EAE models are tasked with extracting arguments like ``government'' and ``shelling'',  which correspond to the roles of \textit{attacker}, and \textit{instrument}.

With the successful application of retrieval-augmented generation (RAG)~\cite{lewis2020retrieval} technology to various NLP tasks~\cite{levonian2023retrieval, li2022survey, ni2024timeseries}, some works~\cite{du2022retrieval, du2022dynamic, ren2023retrieve, huang2023simple} have incorporated retrieval-augmented techniques into event extraction. They use similarity-based retrieval to retrieve the most relevant instances (demonstrations) from the training set for the input query, providing prior external knowledge and augmenting the EAE process.
However, these retrieval-based EAE methods still face some issues that hinder further improvement.

\textbf{First, retrieval augmentation is limited by the model input length.} Current mainstream generation-based EAE approaches typically utilize BART~\cite{lewis2019bart} or T5~\cite{raffel2020exploring} as the PLM. Consequently, due to the input length limitation of these inference models, only a very limited amount of retrieved information can be used for augmentation. For instance, in previous retrieval-based EAE methods~\cite{ren2023retrieve, huang2023simple}, the number of retrieved demonstrations is limited to just one, which significantly limits the diversity of retrieved content.

\textbf{Second, the retrieval quality is limited by the gap between the retriever and the inference model.} Current mainstream retrieval-based EAE methods~\cite{ren2023retrieve, huang2023simple, du2022dynamic} use dense retrievers such as S-BERT~\cite{reimers2019sentence} and retrieve based on the similarity of the context. These retrievers, often untrained, exhibit an embedding gap with inference models as highlighted in recent studies~\cite{ren2022thorough, thakur2021beir, xu2023berm}, leading to sub-optimal retrieval quality.  Additionally, in EAE tasks, only a few contextual words serve as event arguments, while other extraneous content can mislead the retriever, resulting in the retrieval of irrelevant demonstrations. 

%
 Recently, numerous studies~\cite{munkhdalai2024leave, katharopoulos2020transformers, tiezzi2024state, gu2023mamba} have adopted RNN-inspired approaches to tackle the quadratic complexity issue of processing long texts in transformers. Inspired by these works, we propose a \textbf{C}ompressive \textbf{M}emory-based \textbf{R}etrieval (CMR) method for EAE, which effectively addresses the two issues mentioned above. Specifically, we design a compressive memory mechanism that caches the information of retrieved demonstrations. This compressive memory, structured as a dynamic matrix, supports continuous updates and is theoretically capable of caching information indefinitely.
 Before inference, the model pre-loads all candidate demonstrations into the memory. Then it dynamically retrieves necessary information from the memory based on the input query, enabling adaptive filtering of the candidate demonstrations retrieved by the retriever.


Our proposed CMR model have the following two advantages over existing EAE methods: (1) CMR breaks the limitation of the model's context window size, enabling the retrieval of more instances as demonstrations and ensuring the diversity of RAG. (2) CMR enables the model to further filter the retrieved information, reducing the interference from irrelevant information and bridging the gap between the retriever and the inference model. Additionally, we introduce a training strategy that enhances the efficiency of the training process and improves the robustness of the model.
Our contributions are summarized as follows:

\begin{itemize} 
\setlength{\itemsep}{1pt}
\setlength{\parsep}{1pt}
\setlength{\parskip}{0pt}
\item  We propose a Compressive Memory-based Retrieval (CMR) mechanism for EAE, employing a dynamic memory matrix to store retrieved demonstrations. This approach enables existing EAE models to handle larger volumes of retrieved content, significantly enhancing retrieval diversity.
\item Our CMR mechanism can further filter retrieved information from candidate demonstrations, reducing interference from irrelevant information and bridging the gap between the retriever and inference model.

\item Extensive experiments demonstrate that the proposed CMR mechanism outperforms previous retrieved-based EAE methods. Further experimental analysis demonstrates the effectiveness and robustness of our method.

\end{itemize}

\section{Methodology}
In this section, we first provide a formal definition of the EAE task. Consider an instance $\left( X, \{e_i\}^K_{i=1}, \{t_i\}^K_{i=1},  \{R^{\left( e_i \right)} \}^K_{i=1} \right) $, where $X=\{ w_0, w_1, \ldots, w_{N-1} \}$ represents the document text consisting of $N$ words, and $K$ is the number of target events. Here, $e_i$ denotes the type of the $i$-th event, $t_i \subseteq X$ represents the trigger word of the $i$-th event, and $ R^{\left( e_i \right)}$ indicates the set of roles associated with the event $e_i$. The objective is to extract a set of spans $\mathcal{S}_i$ for each event $e_i$, which satisfies $\forall a^{\left( r \right)}\in \mathcal{S}_i,(a^{\left( r \right)}\subseteq X)\land (r\in R^{\left( e_i \right)})$.  Following this, we introduce the traditional RAG architecture for EAE and then describe our proposed Compressive Memory-based Retrieval (CMR) architecture.

\begin{figure*}[htbp]
    \centering
    \includegraphics[width=0.95\linewidth]{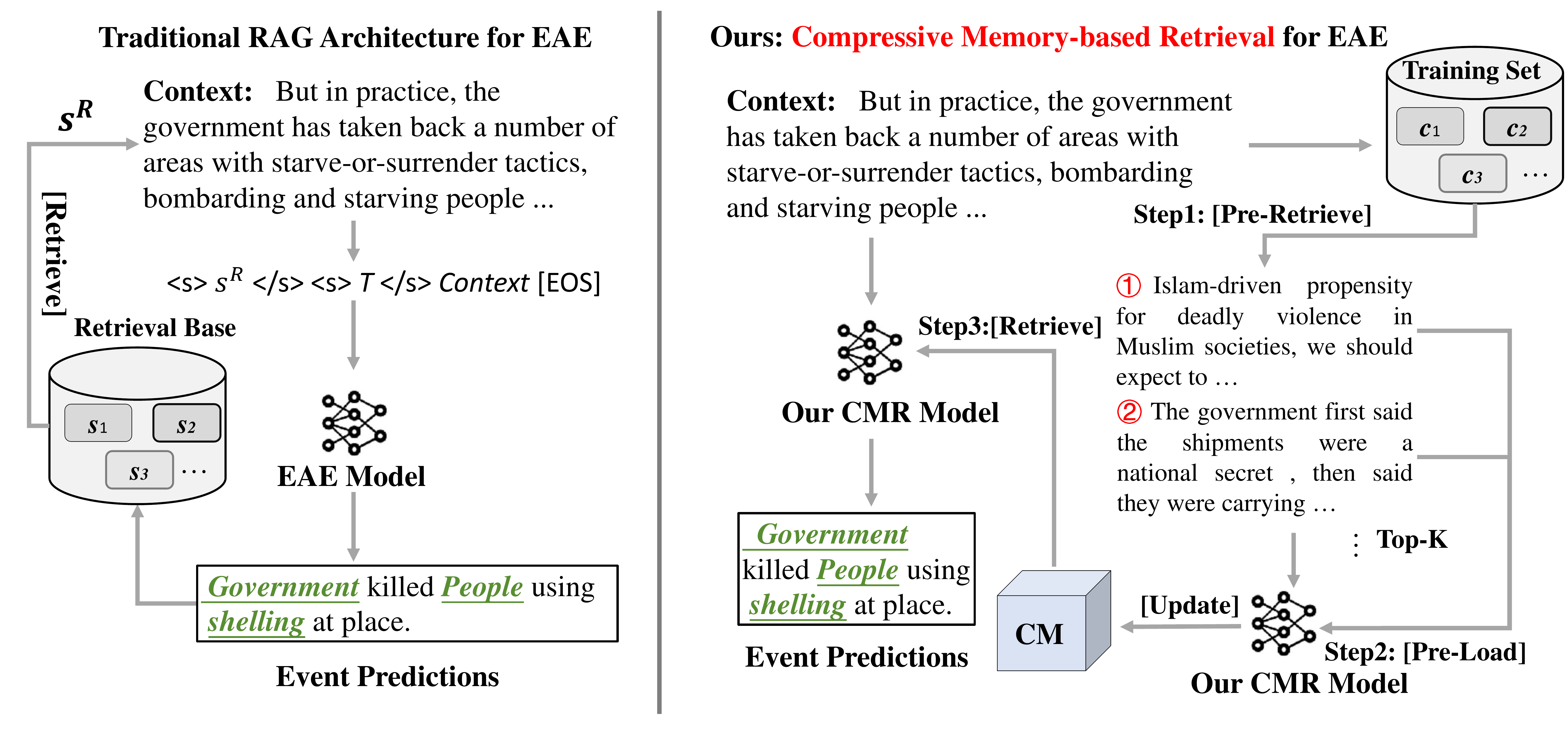}
    
    \caption{Overview of Compressive Memory-based Retrieval architecture. ``CM'' denotes the Compressive Memory. First, the model pre-loads all retrieved candidate demonstrations to build the memory. Then, it dynamically retrieves information from the memory based on the input query, and subsequently generates the final prediction.}
    \label{fig:architecture of model}
\end{figure*}  

\subsection{Traditional RAG Architecture for EAE}
Traditional retrieval-based EAE methods typically retrieve the demonstrations from a knowledge base, such as the training set. Specifically, when predicting the $i$-th event $e_i$ in a document, the knowledge base is $K=\{s_1, s_2, ..., s_n\}$,  where $s_i$ denotes the candidates to be retrieved\footnote{The candidate $s_i$ can be the context~\cite{ren2023retrieve} or event predictions~\cite{du2022dynamic}. In our implementation, we use both the context and event predictions of each instance as candidates (detailed in Section~\ref{sec:3.2}.) }. Then, using S-BERT embeddings~\cite{reimers2019sentence}, the cosine similarity between $e_i$'s context $c_i$ and each candidate in $K$ is calculated, and the candidate with the highest score is selected as additional input to enhance the prediction of $e_i$:
\begin{equation}
\label{equation: eq1}
\setlength\abovedisplayskip{3pt plus 3pt minus 7pt}
\setlength\belowdisplayskip{3pt plus 3pt minus 7pt}
\begin{aligned}
 \text{score}(s_j, c_i) &= \frac{\exp f(c_i, s_j)}{\sum_{s_j \in M} \exp f(c_i, s_j)}, \\[3pt]
f(c_i, s_j) &= \text{S-BERT}(c_i)^T \text{S-BERT}(s_j), \\[3pt]
s_i^R &= \arg\max_{s_j} \text{score}(s_j, c_i),
\end{aligned}
\end{equation}
where $s_i^R$ denotes the retrieved candidate that $e_i$ depends on. Then $s_i^R$ is concatenated as a prefix to the input to enhance the model's performance:
\[\text{Input} = 
\langle s \rangle s_i^R \langle /s \rangle \langle s \rangle P \langle /s \rangle x_1, x_2, \ldots, x_N \text{[EOS].}
\]
where $x_1, x_2, \ldots, x_N$, are the context words, $\langle s \rangle$ 
 and $\langle /s \rangle$ denote special delimiter tokens, and $P$ indicates the task prompt\footnote{Typically, it is an unfilled template~\cite{ma2022prompt, huang2023simple} of the target event.}. $P$ and the context words form the event context $c_i$.

\subsection{Compressive Memory-based Retrieval}
\label{sec:3.2}
Traditional retrieval-based EAE architectures primarily face two major issues: (1) Due to the input length limitation of PLMs, the retrieved content is restricted to the most similar candidate, severely lacking in diversity. (2) The retriever uses fixed parameters and not trained alongside the model to adapt to downstream tasks.

Inspired by Linear Attention mechanism~\cite{katharopoulos2020transformers}, we introduce our Compressive Memory-based Retrieval (CMR) mechanism for EAE in this section.  Our CMR mechanism addresses the above two issues: (1) The CMR mechanism overcomes the limitation of model input length, theoretically enabling the retrieval of an unlimited number of demonstrations. (2) It incorporates a memory retrieval mechanism that can further filter the information, enabling the model to adaptively
retrieve useful information for the EAE task. Utilizing trainable parameters from the PLM, the CMR mechanism effectively bridges the gap between the retriever and the inference model. In Appendix~\ref{appendix:C}, we prove that our CMR mechanism enables the information retrieval of demonstrations stored in memory. 

\noindent \textbf{Compressive Memory.} \quad We design a compressive memory $\mathbf{M}$ for each transformer layer to store candidate demonstrations encountered by the model. Unlike traditional vector retrieval databases, this memory is a fixed-size matrix. Each time the model finishes processing a candidate instance, the memory is updated based on the Key-Value (KV) cache of that instance. 
Note that the compressive memory is not part of the model parameters and can be inserted or removed as needed. When previous memories are no longer required, $\mathbf{M}$ can be reset to zero, effectively erasing all stored information.  

\noindent  \textbf{Memory Storage and Update.} \quad For simplicity, we only illustrate the memory mechanism for a single layer. Given the context of the instance \(q\) and the retrieved demonstrations \(D = \{d_1, d_2, \dots, d_k\}\), our CMR mechanism first stores these demonstrations into the compressive memory. To prevent memory overflow, inspired by~\cite{katharopoulos2020transformers}, we introduce a normalization term \(\mathbf{n} \in \mathbb{R}^{d_k}\), using a sum of all keys for normalization. For each demonstration \(d_i\), represented by the embedding \(\mathbf{X}^{d_i} \in \mathbb{R}^{N \times d_{\text{model}}}\), the memory and normalization term are updated as follows:
 \begin{equation}
\setlength\abovedisplayskip{3pt plus 3pt minus 7pt}
\setlength\belowdisplayskip{3pt plus 3pt minus 7pt}
\mathbf{K}^{d_i} = \mathbf{X}^{d_i}\mathbf{W}_k, \mathbf{V}^{d_i} = \mathbf{X}^{d_i}\mathbf{W}_v, 
\end{equation}
\begin{equation}
\mathbf{M}_i \leftarrow \mathbf{M}_{i-1} + \sigma(\mathbf{K}^{d_i})^T \mathbf{V}^{d_i}, 
\label{eq:mem_update1}
\end{equation}

\begin{equation}
\setlength\abovedisplayskip{3pt plus 3pt minus 7pt}
\setlength\belowdisplayskip{3pt plus 3pt minus 7pt}
\quad \mathbf{n}_i \leftarrow \mathbf{n}_{i-1} + \sum_{j=1}^{N} \sigma(\mathbf{K}^{d_i}_j),
\label{eq:mem_update2}
\end{equation}
where $\mathbf{W}_k \in \mathbb{R}^{d_{\text{model}} \times d_k}$ and $\mathbf{W}_v \in \mathbb{R}^{d_{\text{model}} \times d_v}$ are trainable parameters from the transformer.  Activation function $\sigma$ is the element-wise ELU + 1~\cite{clevert2015fast} function.  

\noindent  \textbf{Memory Retrieval.} \quad The process of memory retrieval is integrated into the transformer's multi-head attention mechanism. For the instance $q$, represented by the embedding $\mathbf{X} \in \mathbb{R}^{N \times d_{\text{model}}}$, we initially calculate the vanilla dot-product attention (for a single head) $\mathbf{A}_\text{dot} \in \mathbb{R}^{N \times d_{v}}$ as follows:
\begin{equation}
\mathbf{A}_{\text{dot}} = \text{softmax}\left(\frac{\mathbf{Q}\mathbf{K}^T}{\sqrt{d_{\text{model}}}}\right) \mathbf{V},
\end{equation}
\begin{equation}
\mathbf{K} = \mathbf{X}\mathbf{W}_k, \mathbf{V} = \mathbf{X}\mathbf{W}_v, \mathbf{Q} = \mathbf{X}\mathbf{W}_q.
\end{equation}

Subsequently, we utilize the input query $\mathbf{Q} \in \mathbb{R}^{N \times d_{k}}$ to retrieve from memory, obtaining the retrieval-augmented representation $\mathbf{A}_{\text{ret}}\in \mathbb{R}^{N \times d_{v}}$:

\begin{equation}
\mathbf{A}_{\text{ret}} = \frac{\sigma(\mathbf{Q}) \mathbf{M}_{k}}{\sigma(\mathbf{Q}) \mathbf{n}_{k}}.
\label{equation:retrieval}
\end{equation}
Here, $\mathbf{M}_{k} \in \mathbb{R}^{d_k \times d_{v}}$ is the compressive memory that stores information of all demonstrations, and $\mathbf{n}_{k} \in \mathbb{R}^{d_{k}}$ is the normalization term, which is crucial for training stability.   

Then, we combine the vanilla dot-product attention $\mathbf{A}_\text{dot}$ and the retrieved $\mathbf{A}_\text{ret}$ using a gating mechanism:

\begin{equation}
\setlength\abovedisplayskip{3pt plus 3pt minus 7pt}
    \mathbf{A} = \textit{S}(\gamma) \odot \mathbf{A}_{\text{ret}} + (1 - \textit{S}(\gamma)) \odot \mathbf{A}_{\text{dot}},
\end{equation}
where $\gamma$ is a trainable gating scalar, and $\textit{S}(\cdot)$ denotes the Sigmoid function. Through the trainable gating scalar $\gamma$, the model achieves a learnable balance between input and retrieved information.
Note that since the stored KV entries implicitly include the model's predictions, our memory update process retains both the context of candidate demonstrations and the model's event predictions.


\subsection{Implementation}
The proposed CMR mechanism can be well applied to both encoder-decoder and decoder-only architectures. (1) For models with an encoder-decoder architecture, the operations described in Section~\ref{sec:3.2} are implemented in the cross-attention module of each decoder layer, using the decoder's input as $\mathbf{Q}$ illustrated in Equation \ref{equation:retrieval}. (2) For decoder-only models, we replace the vanilla self-attention mechanism in each layer with our CMR mechanism.

\subsubsection{Training} 
\label{sec:training}
During the training process, we need to teach the model how to retrieve relevant information from memory to enhance generation for the EAE task. However, pre-retrieving the top-k-related candidate demonstrations for each training instance entails certain limitations: (1) The fixed number of retrieved demonstrations during training may restrict the model to a specific demonstration count, limiting the roubstness of RAG. (2) Such a training approach is very time-consuming.

Therefore, we propose an efficient and robust training method. Specifically, we set a maximum retrieval number $\textit{Max\_{retrieval}}$ and initialize the memory $\mathbf{M}_0$ to zero.  Within $\textit{Max\_{retrieval}}$, the model updates its memory as it infers each training instance\footnote{These stored instances will act as demonstrations for subsequent training instances.}.  When the number of instances stored in memory exceeds $\textit{Max\_{retrieval}}$, the memory is reset to zero and the cycle repeats. The $\textit{Max\_{retrieval}}$ is set to match the model's gradient accumulation steps. 
To ensure the relevance of the retrieved information, we rerank the shuffled training data in each epoch, organizing batches so that each training instance is primarily surrounded by instances of the same event type\footnote{In EAE task, instances of the same event type often have high relevance to each other~\cite{ebner2020multi, huang2023simple}.}, while also including a strategic mix of instances from different types to enhance model generalization and prevent overfitting. The detailed training algorithm is outlined in Algorithm~\ref{alg:alg1} in Appendix~\ref{appendix:1}.

Our proposed training method has the following two advantages: (1) It significantly reduces training time. (2) Within each $\textit{Max\_{retrieval}}$, the count of instances stored in memory continuously increases. This naturally provides training instances with varying retrieval numbers, which enables the model to adapt to varying retrieval volumes, enhancing its robustness.

\begin{table*}[htbp]
\centering

\small{
\setlength{\tabcolsep}{1.5mm}{
\begin{tabular}{lllllllll}
\hline
\multirow{2}{*}{Scheme} 
& \multirow{2}{*}{Method}  
& \multirow{2}{*}{PLM}
& \multicolumn{2}{c}{RAMS} & \multicolumn{2}{c}{WikiEvents} & \multicolumn{2}{c}{ACE2005} \\ \cline{4-9} 
                                 &   &    & Arg-I & Arg-C & Arg-I & Arg-C & Arg-I & Arg-C \\ \hline
                               
\multirow{6}{*}{\textit{W.o. Retrieval}} 

& ${\text{DEEIA}}$~(\citeyear{liu2024beyond}) & ${\text{RoBERTa-l}}$       &  \underline{58.0} & \underline{53.4} & \underline{71.8} & \underline{67.0} & 76.3      & 74.1      \\
& ${\text{TabEAE}}$~(\citeyear{he2023revisiting})  & ${\text{RoBERTa-l}}$ & {57.3}     & {52.7}     & 71.4 & 66.5     &   \textbf{77.2}     &   \textbf{75.0}     \\
& ${\text{SPEAE}}$~(\citeyear{nguyen2023contextualized})  & ${\text{BART-l}}$ & {58.0}     & {53.3}     & 71.9 & 66.1     &   \quad -     &    \quad -     \\
& ${\text{SCPRG}}$~(\citeyear{liu2023enhancing}) & ${\text{RoBERTa-l}}$   & 56.7  & 52.3  &71.3  & {66.4}  &  \quad -     & \quad -      \\ 

& ${\text{PAIE}}$~(\citeyear{ma2022prompt})   & ${\text{BART-l}}$ & 56.8     & 52.2     & 70.5  & 65.3     &  72.1     &  70.8     \\

& ${\text{BART-Gen}}$~(\citeyear{lietal2021document}) & ${\text{BART-l}}$ & 51.2     & 48.6     & 66.8  & 62.4 & 69.9      & 66.7      \\ 

 \hline

\multirow{6}{*}{\textit{With Retrieval}}  
& ${\text{R-GQA}}$~(\citeyear{du2022retrieval})  & ${\text{BART-l}}$    & \quad - & \quad - & \quad - & \quad- &  75.5       &   72.8     \\ 
& ${\text{AHR}}$~(\citeyear{ren2023retrieve})  & ${\text{T5-l}}$    &  54.6 & 48.4 & 69.6 & 63.4 &  \quad-       &   \quad-    \\ 

 
& ${\text{PAIE-R}}$* & ${\text{BART-l}}$ & 57.4 & 53.0 & 71.2 & 66.0 & 73.0 & 71.9 \\ 
& ${\text{BART-Gen-R}}$* & ${\text{BART-l}}$ & 51.4 & 49.1 & 67.9 & 63.2 & 70.2 & 66.9 \\ 
\rowcolor{gray!20}
& ${\text{PAIE-CMR (Ours)}}$ & ${\text{BART-l}}$ & \textbf{59.1} \tiny{($\uparrow$ 1.7)} & \textbf{54.3} \tiny{($\uparrow$1.3)} & \textbf{72.8} \tiny{($\uparrow$1.6)} & \textbf{67.9} \tiny{($\uparrow$1.9)} & \underline{76.8} \tiny{($\uparrow$3.8)} & \underline{74.8} \tiny{($\uparrow$2.9)} \\ 
\rowcolor{gray!20}
& ${\text{BART-Gen-CMR (Ours)}}$ & ${\text{BART-l}}$ & 53.2 \tiny{($\uparrow$1.8)} & 51.4 \tiny{($\uparrow$2.3)} & 69.1 \tiny{($\uparrow$1.2)} & 65.3 \tiny{($\uparrow$2.1)} & 72.4 \tiny{($\uparrow$2.2)} & 69.3 \tiny{($\uparrow$2.4)} \\ 
\hline

\end{tabular}
}
\caption{
Comparison of performance on RAMS, WikiEvents, and ACE2005 test set. * means that we add vanilla retrieval into the original method. 
The shaded area represents our methods, which retrieve top-10 demonstrations.
\textbf{Bold} and \underline{underline} indicate the best and second-best experimental results. 
}

\label{tab:main_results}
}
\end{table*}

  \subsubsection{Inference} 
  During inference, the model first pre-loads all candidate demonstrations to build memory. Specifically, each retrieved demonstration (context) is fed into the model, and the memory is updated according to Equations \ref{eq:mem_update1} and \ref{eq:mem_update2}. Notably, during the pre-loading of each demonstration, the memory is only updated but does not participate in the attention calculation. To improve efficiency, we pre-load candidate demonstrations in batches, significantly reducing inference time. 
  
  Subsequently, the model dynamically retrieves necessary information from the memory based on the input query (context of the current inference instance), facilitating adaptive filtering of information from candidate demonstrations.  As for the input order of candidate demonstrations, we illustrate in the experimental section that our model is not sensitive to the input order. 
  The inference algorithm is detailed in Algorithm \ref{alg:alg2} in Appendix~\ref{appendix:1}. 

\begin{figure*}[t]
  \centering
\begin{subfigure}{0.325\linewidth}
    \centering
    \includegraphics[width=\linewidth]{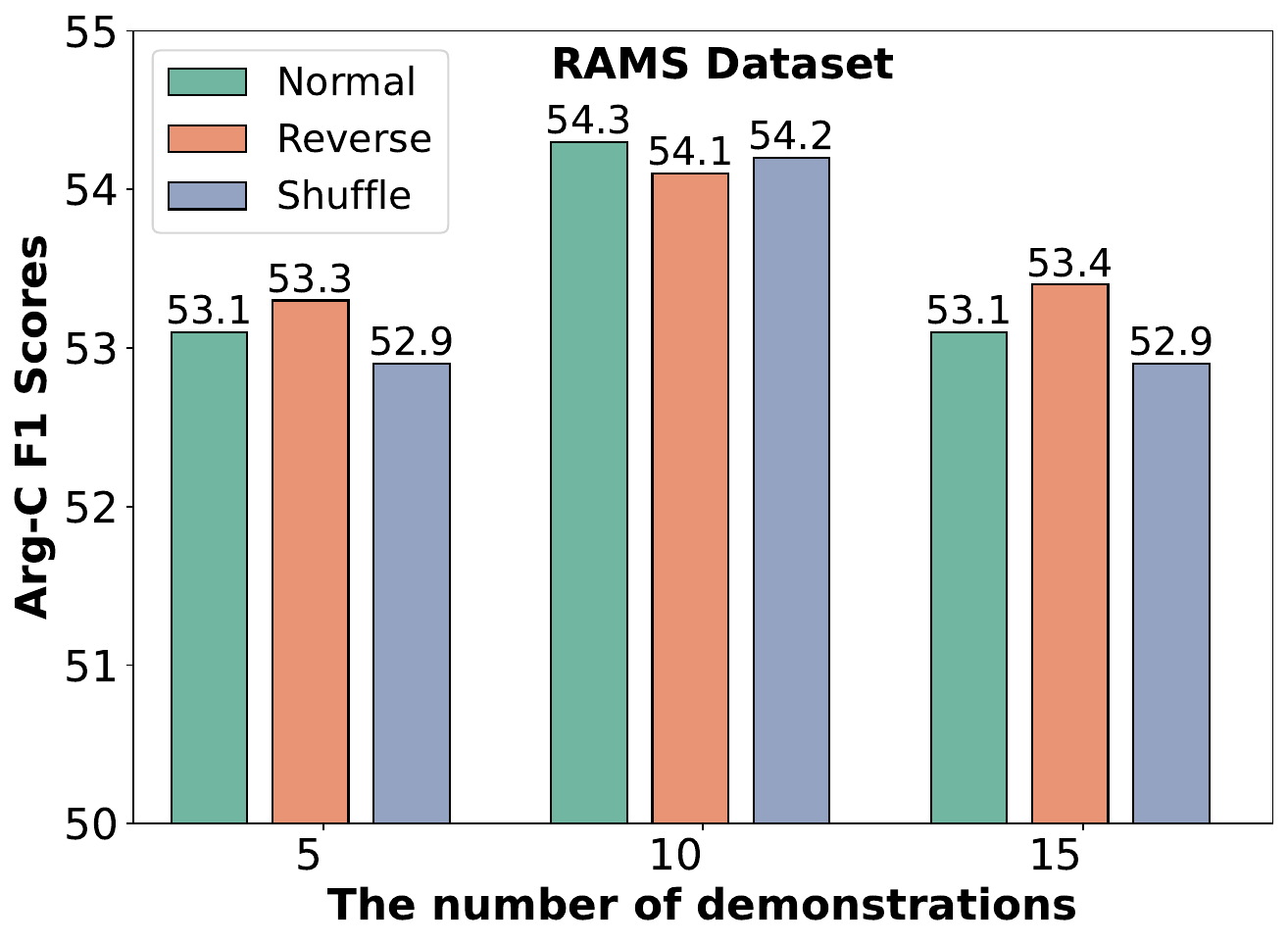}
    
    \label{fig:number_demo_rams}
  \end{subfigure}
  \begin{subfigure}{0.325\linewidth}
    \centering
    \includegraphics[width=\linewidth]{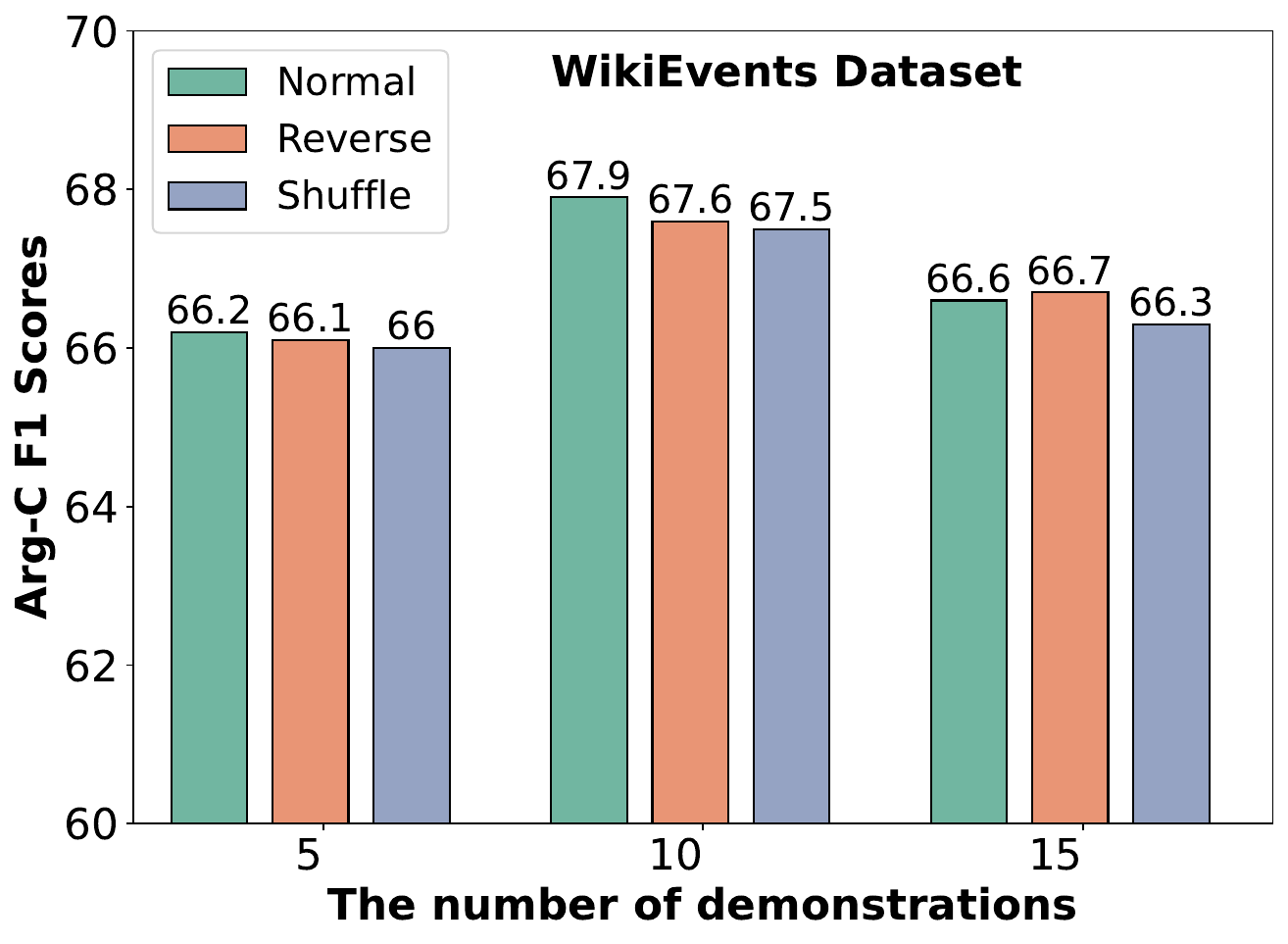}

    \label{fig:number_demo_wiki}
  \end{subfigure}
  \begin{subfigure}{0.325\linewidth}
    \centering
    \includegraphics[width=\linewidth]{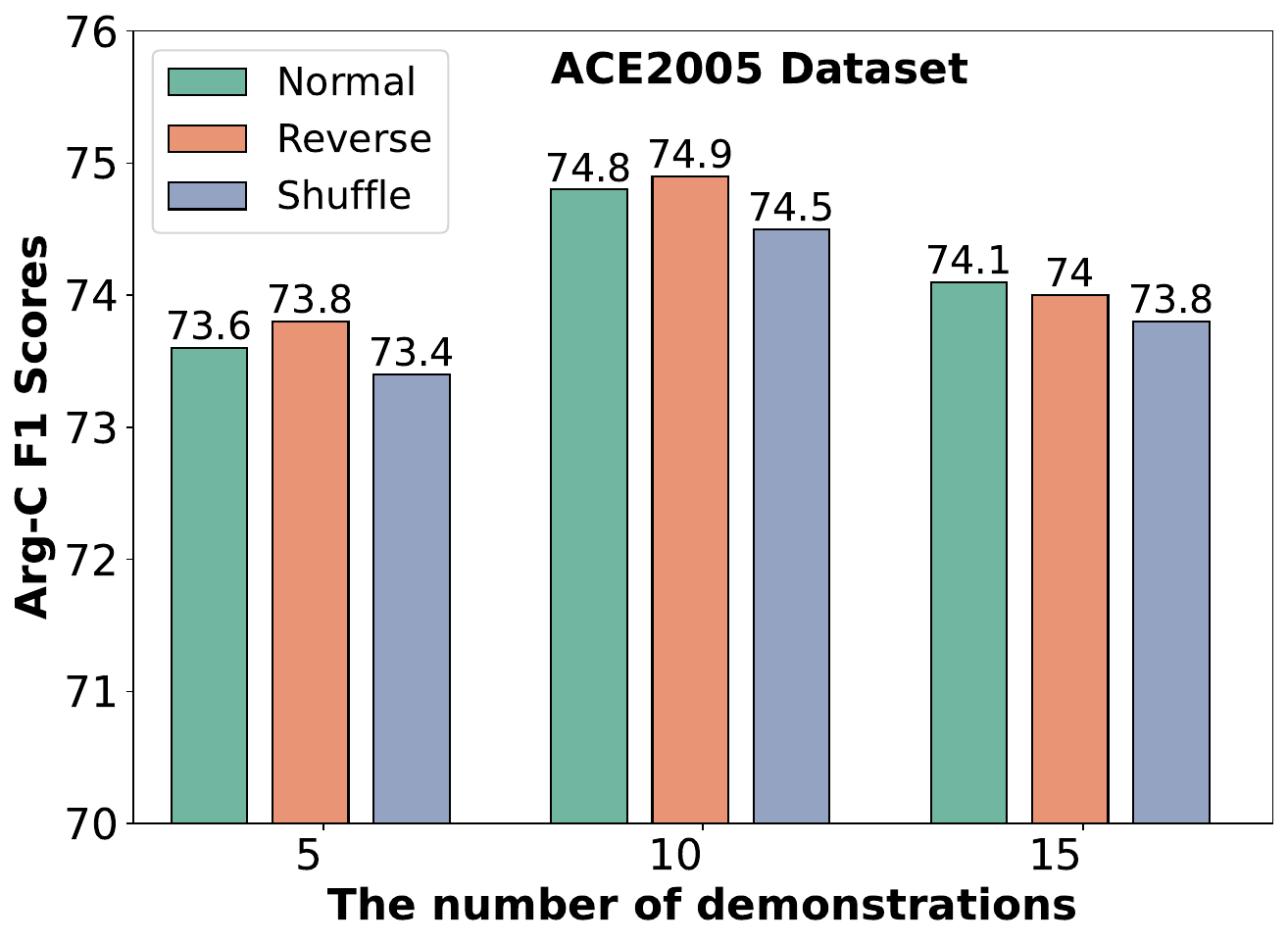}

    \label{fig:number_demo_ace}
  \end{subfigure}
  \caption{Demonstrations order experiment for PAIE-CMR. \textbf{Normal} uses the top-k demonstrations in their original retrieved order, \textbf{Reverse} uses them in the opposite order, and \textbf{Shuffle} means randomly shuffling the demonstrations.}
  \label{fig:demo order}
\end{figure*}


\begin{table}[ht]
\centering

\small{
\setlength{\tabcolsep}{0.8mm}{
\begin{tabular}{cccccccc}
\hline
\multirow{2}{*}{Method} 
& \multirow{2}{*}{\#N }  
& \multicolumn{2}{c}{RAMS} & \multicolumn{2}{c}{WikiEvents} & \multicolumn{2}{c}{ACE2005} \\ \cline{3-8} 
                                 &    & Arg-I & Arg-C & Arg-I & Arg-C & Arg-I & Arg-C \\ \hline
                                 \multirow{5}{*}{\text{PAIE-CMR}} 
& 0  & {56.1}  & {51.8} & {70.6} & {64.8}  & {71.9}  & {70.4}  \\       & 1  & {57.8}  & {53.1} & {71.4} & {66.2}  & {74.3}  & {72.9}  \\ 
& 5  & {58.5}  & {53.6} & {72.0} & {66.9}  & {75.2}  & {73.6}  \\ 
& 10  & \textbf{59.1}  & \textbf{54.3} & \textbf{72.8} & \textbf{67.9}  & \textbf{76.8}  & \textbf{74.8}  \\ 
& 15  & {58.8}  & {54.0} & {72.4} & {67.5}  & {76.2}  & {74.1}  \\ 
\hline

                                 \multirow{5}{*}{\text{B-G-CMR}} 
& 0  & {51.0}  & {47.5} & {66.0} & {61.7}  & {69.2}  & {66.4}  \\ 
& 1  & {52.0}  & {49.9} & {68.4} & {64.1}  & {70.8}  & {67.7}  \\ 
& 5  & {52.7}  & {50.8} & \textbf{69.1} & \textbf{65.3}  & {72.0}  & {69.1}  \\ 
& 10  & \textbf{53.2}  & \textbf{51.4}   & {68.9} & {64.7} & \textbf{72.4}  & \textbf{69.3}  \\ 
& 15  & {52.9}  & {51.0} & {68.5} & {64.1}  & {71.7}  & {68.6}  \\ 
\hline
\end{tabular}
}
\caption{
The performance of retrieving varying numbers of demonstrations (only context) in CMR mechanism. \#N  represents the number of retrieved top-k demonstrations, with \#N  equals to 0 indicating no retrieval.}

\label{tab:different number retrievals}
}

\end{table}

\begin{table*}[ht]
\centering
\small{
\setlength{\tabcolsep}{3mm}{
\begin{tabular}{cccccccc}
\hline
\multirow{2}{*}{Method} & \multirow{2}{*}{\#N } & \multicolumn{2}{c}{RAMS}                  & \multicolumn{2}{c}{WikiEvents}             & \multicolumn{2}{c}{ACE2005}                \\ \cline{3-8} 
                        &                      & Strict F1           & Relaxed F1          & Strict F1            & Relaxed F1          & Strict F1           & Relaxed F1           \\ \hline
LLaMA3-SFT-CMR          & 5                    & 32.48        & 35.25         & 23.11        & 31.63           & 30.72       & 42.97         \\
LLaMA3-SFT-CMR          & 10                   & 32.78 & \text{37.24} & \textbf{24.22} & \textbf{32.96} & \textbf{31.21} & \textbf{43.54} \\ \hline
LLaMA3-SFT              & 0                    & 31.05           & 34.93            & 22.41           & 30.99          & 27.81           & 40.28         \\

LLaMA3-SFT              & 5                    & 31.65         & 35.63             & 22.77          & 31.00         & 28.88         & 41.69          \\
LLaMA3-SFT              & 10                   & \text{32.96}         & 36.10         & 21.23          & 30.64         & 29.49         & 41.35          \\ \hline
LLaMA3-SFT-R              & 0                    & 29.74          &  32.68           &   19.43       &    29.67       &   28.95       &  39.40         \\
LLaMA3-SFT-R              & 5                   & 32.72         &  35.34        &  22.02          &  31.08        &  30.46        & 41.66           \\ 
LLaMA3-SFT-R              & 10                    & \textbf{33.08}          &  \textbf{37.42}            &   22.80       &    31.63       &   30.28       &  41.87         \\ \hline

\end{tabular}
\caption{
Performance comparison of models fine-tuned on LLaMA3-8b-instruct. LLaMA3-8b-SFT, LLaMA3-8b-SFT and LLaMA3-8b-SFT-R are all trained on the RAMS training set and then evaluated on the RAMS, WikiEvents, and ACE2005 test sets. \#N  indicates the number of retrieved demonstrations  (only context) from corresponding training set. \textbf{Bold} highlights the best experimental results.
}
\label{tab:LLM results} }
}
\end{table*}
\section{Experiments}
This section applies the proposed CMR mechanism to the current mainstream EAE baselines across three commonly used EAE benchmarks. Subsequently, we extend the CMR mechanism to decoder-only large language models to further explore its effectiveness. Additionally, we conduct detailed analytical experiments to analyze our method across various settings.

\subsection{Experimental  Setup}
\subsubsection{Datasets}  We conduct experiments on three widely used EAE datasets: RAMS~\cite{ebner2020multi}, WikiEvents~\cite{lietal2021document}, and ACE2005~\cite{doddington2004automatic}. Detailed descriptions of these datasets are provided in Appendix~\ref{appendix:data_statistic}.

\subsubsection{Baselines} We categorize the baselines for comparison into two groups: \textbf{W.o. Retrieval} and \textbf{With Retrieval}. 

\noindent \textbf{W.o. Retrieval:}  We select recent state-of-the-art EAE methods, including DEEIA~\cite{liu2024beyond}, TabEAE~\cite{he2023revisiting}, SPEAE~\cite{nguyen2023contextualized}, SCPRG~\cite{liu2023enhancing}, PAIE~\cite{ma2022prompt}, and BART-Gen~\cite{lietal2021document}. 

\noindent \textbf{With Retrieval:} We choose some classic retrieval-based EAE methods, including R-GQA~\cite{du2022retrieval} and AHR~\cite{ren2023retrieve}. Since previous retrieval-based EAE methods did not use uniform datasets and metrics for evaluation, to ensure a more comprehensive and fair comparison, we establish two retrieval-based EAE baselines PAIE-R and BART-Gen-R based on two commonly used methods, PAIE and BART-Gen. 
Specifically, we follow~\cite{du2022retrieval}, using the S-BERT retriever to identify and incorporate the most relevant (Top-1) event prediction as a prefix into the input.

\subsubsection{Evaluation Metrics}
Following earlier studies~\cite{ma2022prompt, he2023revisiting}, we evaluate the performance using two metrics: (1) Argument Identification F1 (Arg-I), which deems a predicted event argument correct if its boundaries align with any corresponding reference arguments. (2) Argument Classification F1 (Arg-C), requiring both boundary and role type accuracy for a predicted event argument to be considered correct. Our experiments are conducted five times with different seeds, and we report the average results.


\subsection{Main Results}
\noindent \textbf{Comparison with W.o. Retrieval methods.}  As shown in Table~\ref{tab:main_results}, our PAIE-CMR and BART-Gen-CMR models outperform previous non-retrieval SOTA methods, such as SCPRG and DEEIA, showcasing a strong competitive advantage.

\noindent \textbf{Comparison with  Retrieval-based methods.} As shown in Table~\ref{tab:main_results}, two classic EAE baselines, PAIE and BART-Gen, achieve improved performance across all three datasets after incorporating retrieval, which highlights the positive impact of RAG on the EAE task. However, the performance improvement of PAIE-R and BART-Gen-R over the baseline is minimal, demonstrating the limitations of previous retrieval-based EAE methods.   These methods are restricted to retrieving only the top-1 demonstration, which severely lacks diversity and results in sub-optimal performance. In contrast, our CMR mechanism ensures the diversity of retrieved demonstrations and further filters the information, achieving superior performance.
\subsection{CMR for Decoder-Only LLMs}
In this section, we explore the effectiveness of our CMR mechanism on decoder-only LLMs. We fine-tune LLaMA3-8b-instruct~\cite{touvron2023open} on the RAMS dataset and evaluate the performance of our method.

\noindent \textbf{Evaluation Metrics.} \quad
We establish two evaluation metrics to evaluate the performance of the LLM-based EAE models: (1) \textbf{Strict-F1}, which considers a predicted event argument correct if the model’s prediction exactly matches the golden label. (2) \textbf{Relaxed-F1}, which considers a prediction correct if the golden label is contained within the model’s prediction.

\noindent \textbf{Experimental Details.} \quad
We select LLaMA3-8b-instruct for full-parameter fine-tuning on RAMS training set and evaluate it on the RAMS, WikiEvents, and ACE2005 test sets. First,  we train LLaMA3-SFT-CMR using the CMR mechanism, following the training strategy outlined in Section~\ref{sec:training}. For comparison, we also train a LLaMA3-SFT model using standard supervised fine-tuning. The inference process follows Algorithm~\ref{alg:alg2}. Additional training details, including prompts and experimental settings, are provided in Appendix~\ref{sec：appendix_decoder}.

\noindent \textbf{Analysis.} \quad As shown in Table~\ref{tab:LLM results}: (1) For LLaMA3-SFT, the impact of RAG after supervised fine-tuning is minimal, with some cases even showing a decline in performance. (2) In contrast, our LLaMA3-SFT-CMR model performs better when retrieving more demonstrations, underscoring the effectiveness of our CMR mechanism in decoder-only LLM architectures and demonstrating the generalizability of our approach. (3) However, the overall improvement of LLaMA3-SFT-CMR over LLaMA3-SFT remains limited. We assume that this is due to the large parameter size of the LLaMA3-8b-instruct model, combined with the relatively small size and limited task diversity of the fine-tuning data, which may hinder the model's ability to fully learn the CMR capability.
\section{Analysis}
In this section, we further analyze our CMR mechanism by addressing the following questions: Q1: How does the CMR mechanism compare to directly using a long-context model? Q2: How does the number of demonstrations during inference affect performance? Q3: What impact does the order of demonstrations have on performance? Q4: Can this method filter out irrelevant information and enhance the robustness of the RAG?
\subsection{Q1: Compare with Long-Context Models}
To evaluate the effectiveness of the CMR mechanism compared to directly using a long-context model, we select LLaMA3-8b-instruct as the base model and train \textbf{LLaMA3-SFT-R} model through retrieval-based training. Aligning with the the training process of our CMR mechanism, we retrieve top 8 demonstrations for each training instance and insert these demonstrations into the prompt in Figure~\ref{fig:prompt}. The remaining fine-tuning details are consistent with those of LLaMA3-SFT.  

As shown in Table~\ref{tab:LLM results},  LLaMA3-SFT-R significantly improves performance over the non-retrieval scenario with retrieval. Additionally, although LLaMA3-SFT-R performs well on the RAMS dataset, it generalizes poorly to  WikiEvents and ACE2005  when compared to our LLaMA3-SFT-CMR model. This suggests that simply using a long-context model to directly train RAG capabilities for EAE results in poor generalization. In contrast, our model learns to adaptively retrieve and filter information from memory during training, which enhances the generalization capability.

\subsection{Q2: Analysis on Demonstration Numbers}
Table~\ref{tab:LLM results} shows the performance of PAIE-CMR and BART-Gen-CMR across different numbers of demonstrations.
(1) When \#N  is 1, our CMR approach outperforms PAIE-R and BART-Gen-R. This improvement can be attributed to two reasons: (a) Our method uses more comprehensive demonstrations, including both context and implicit event predictions. 
(b) Our CMR mechanism adaptively filters retrieved information, reducing interference from irrelevant data.
(2) As \#N  increases, the performance shows an improving trend across all three datasets. It suggests that the growing amount and diversity of retrieved information contributes to enhanced performance. Furthermore, it demonstrates that our CMR mechanism effectively stores information from candidate demonstrations and retrieves useful information efficiently. (3) However, when \#N  exceeds $10$, the performance declines. We attribute this to the number of retrieved demonstrations surpassing the training limit of $\textit{Max\_{retrieval}}$, making it difficult for the model to effectively store and manage the excessive information.

\begin{table}[]
\small{
\setlength{\tabcolsep}{1.2mm}{
\begin{tabular}{ccccccc}
\hline
\multirow{2}{*}{Method}   & \multirow{2}{*}{\#N } & \multirow{2}{*}{\#Mode} & \multicolumn{2}{c}{RAMS} & \multicolumn{2}{c}{WikiEvent} \\ \cline{4-7} 
                          &                      &                       & Arg-I       & Arg-C      & Arg-I         & Arg-C         \\ \hline
PAIE                      & 0                    & No Ret.          & 56.8        & 52.2       & 70.5          & 65.3          \\ \hline
\multirow{2}{*}{PAIE-R}   & 1                    & Top-k         & 57.4        & 53.0       & 71.2          & 66.0          \\
                          & 1                    & Random       & 56.2        & 51.5       & 70.1          & 64.4          \\ \hline
\multirow{4}{*}{PAIE-CMR} & 1                    & Top-k         & 57.6        & 53.1       & 71.4          & 66.2          \\
                          & 1                    & Random       & 57.2        & 52.5       & 70.6          & 65.6          \\
                          & 5                    & Top-k         & 58.5        & 53.6       & 72.0          & 66.9          \\
                          & 5                    & Random       & 57.7        & 53.1       & 71.4          & 66.6          \\ \hline
\end{tabular}
}}
\caption{ 
Experiments on retrieval robustness. We compare PAIE-R with our PAIE-CMR, highlighting the robustness of our retrieval method. \#Mode=\{No Retrieval, Top-k Retrieval, Random Retrieval\} represents the different retrieval modes. Random retrieval involves randomly selecting demonstrations from the training set. 
}

\label{tab: random or topk}
\end{table}

\subsection{Q3: Analysis on Demonstration Order}

To explore our method's sensitivity to the order of demonstrations, we design three types of input orders—\textbf{Normal}, \textbf{Reverse}, and \textbf{Shuffle}—and conduct inference on trained checkpoints from three datasets, respectively. We first retrieve the top-k demonstrations and then conduct inference using the PAIE-CMR in the aforementioned three orders. As illustrated in Figure~\ref{tab: random or topk}, when the number of demonstrations is held constant, the performance of the three orders exhibits negligible variation, indicating that our method is insensitive to the order of demonstrations. We assume this is due to the shuffling of instances during training across each epoch, which makes the memory mechanism insensitive to the order of demonstrations, significantly enhancing the robustness of our model.

\subsection{Q4: Retrieval Robustness Analysis}
To explore the retrieval robustness of our method, we implement two retrieval strategies: (1) \textbf{Topk}, which retrieves the top-k most similar demonstrations. (2) \textbf{Random}, which selects demonstrations randomly from the training set. As shown in Table~\ref{tab: random or topk}, the traditional retrieval-based EAE method, PAIE-R, is highly sensitive to the relevance of the retrieved content. Its performance declines significantly with random retrieval, even dropping below that of using no retrieval at all. In contrast, our CMR mechanism demonstrates stronger robustness under conditions of random retrieval. This robustness is attributed to our training strategy, where we maintain a selection of unrelated demonstrations in memory during each gradient update. This strategy significantly enhances the robustness of our model's retrieval-augmented generation.
Furthermore, our CMR mechanism adaptively filters out irrelevant information, effectively reducing interference from noisy data.

In Appendix~\ref{appendix:domain_transfer}, we also conduct experiments to evaluate our model's performance with RAG under new ontologies, demonstrating its robust generalizability across domain transfer scenarios.


\section{Related Works}

\subsection{Event Argument Extraction}
Event argument extraction (EAE) aims to extract specific details about the identified events, such as their locations or the individuals involved, which is a challenging subtask of event extraction.
Recent mainstream EAE methods can be primarily divided into following two categories. (1) Span-based methods, which identify candidate spans and predict their roles~\cite{zhang2020two, yang2023amr, liuetal2017exploiting, zhang2020two, liu2023enhancing, xu2022two}. (2) Generation-based methods, which have recently gained popularity, utilize slotted templates and a generative slot-filling strategy for argument extraction~\cite{ma2022prompt, he2023revisiting, nguyen2023contextualized,lietal2021document,huang2023simple, zeng2022ea}. While both methods offer distinct advantages, generation-based methods have demonstrated superior generalizability and competitive performance compared to their span-based counterparts~\cite{hsu2023ampere}.  

With the advancement of RAG technology~\cite{lewis2020retrieval}, some works~\cite{du2022retrieval, ren2023retrieve, huang2023simple} have incorporated RAG techniques into event extraction, leading to some performance boost. 
However, these methods are constrained by the model's input length, resulting in a limited amount of content available for retrieval enhancement, which significantly restricts both the diversity and quality of RAG. These methods also suffer from a substantial information gap between the retriever and the inference model, which leads to sub-optimal performance.

\subsection{RNN-Inspired Memory Methods for Transformers}
Recently, numerous studies have adopted RNN-inspired approaches to tackle the quadratic complexity issue of processing long texts in transformers. For example, \cite{katharopoulos2020transformers} introduces Linear Attention, which reduces complexity by efficiently retaining relevant information. Similarly, \cite{munkhdalai2024leave} proposes the Infinite Transformer, which utilizes the memory mechanism to allowing the model to focus on previously stored information. Additionally, Mamba~\cite{gu2023mamba} incorporates memory-augmented attention, storing crucial past information for future reference. \cite{tiezzi2024state} leverages state-space models to manage long-range dependencies.  Inspired by these works, we propose a compressive memory mechanism that adaptively retrieves and dynamically updates stored information.
\section{Conclusion}

In this paper, to address the limitations of input length constraints and the gap between the retriever and inference model in existing retrieval-based EAE methods, we introduce a Compressive Memory-based Retrieval  mechanism for EAE. Our approach leverages a dynamic, continuously updating matrix to efficiently cache and manage retrieved information. By pre-loading candidate demonstrations and dynamically filtering based on the input query, our model significantly enhances retrieval quality. Extensive experiments on three public datasets demonstrate that our method achieves new state-of-the-art performance, outperforming existing retrieval-based EAE methods.

\section{Limitations}
The improvement of our CMR mechanism when applied to LLM models like LLaMA3-8b-instruct is limited. We assume this is due to the large number of model parameters combined with the relatively small scale and limited diversity of our training data. Additionally, previous studies have demonstrated the effectiveness of linear attention mechanisms in LLMs~\cite{munkhdalai2024leave, katharopoulos2020transformers}. We plan to explore this further in the future, aiming to extend our CMR mechanism to a broader range of NLP tasks, including generative tasks, such as question answering.
\bibliography{custom}

\clearpage
\appendix

\section{Training and Inference Details}
\label{appendix:1}
We propose an efficient and robust training method, and the detailed  algorithm is shown in Algorithm~\ref{alg:alg1}. For clarity, we only describe the memory update process. Details on normalization and other operations can be found in Section~\ref{sec:3.2} of the main text.
 The \texttt{ShuffleRerank} function first shuffles the training data to eliminate sequence-based patterns, promoting model generalization. After shuffling, the data is reranked by event type, ensuring each batch primarily contains instances of the same event type, with a strategic mix of 20\% different types included to further enhance generalization and prevent overfitting. In the training process, data within each batch is processed in parallel. 

\begin{algorithm}
\caption{Efficient Training of CMR}
\begin{algorithmic}[1]
\REQUIRE Training data $T = \{s_1, s_2, \dots, s_n\}$, Maximum retrieval number $\textit{Max\_{retrieval}}$, Model $\mathcal{M}$
\ENSURE Trained model $\mathcal{M}$
\STATE $\mathbf{M}_0 \leftarrow \mathbf{0}$, $t \leftarrow 1$
\FOR{epoch $e = 1$ to $E$}
    \STATE $\mathbf{D}_e \leftarrow \text{ShuffleRerank}(T)$ // {Shuffle and rerank by event type}
    \FOR{batch $b \subset \mathbf{D}_e$}
        \FOR{instance $s_i \in b$}
            \STATE $\mathbf{O}_t, \mathbf{M}_t \leftarrow \mathcal{M}(\mathbf{M}_{t-1}, s_i)$ // {Forward propagate, $\mathbf{O}_t$ denotes the event predictions of the model.}
            \STATE $t \leftarrow t + 1$
        \ENDFOR
        \STATE $\mathbf{M}_t \leftarrow \mathbf{M}_{t-|b|} + \frac{1}{|b|} \sum_{i=1}^{|b|} \mathbf{M}_{t-|b|+i}$ // {Update memory}
        \IF{$t > \textit{Max\_{retrieval}}$}
                \STATE $\mathbf{M}_0 \leftarrow \mathbf{0}$, $t \leftarrow 1$ // {Reset memory and counter}
                \STATE Update model parameters of $\mathcal{M}$
            \ENDIF
        
    \ENDFOR
\ENDFOR
\end{algorithmic}
\label{alg:alg1}
\end{algorithm}

The detailed inference process is shown in Algorithm~\ref{alg:alg2}. \texttt{RetrieveTopK} uses S-BERT to retrieve the top-k relevant demonstration contexts based on similarity. During inference, data within each demonstration batch $B_j$ is processed in parallel (as seen in lines 4-6 of Algorithm~\ref{alg:alg2}), significantly improving inference efficiency. 

\begin{algorithm}
\caption{Inference with CMR}
\begin{algorithmic}[1]
\REQUIRE Knowledge base $K$, Input query $q$, Model $\mathcal{M}$, Retrieval number $k$
\ENSURE Inference result for query $q$
\STATE $D \leftarrow \text{RetrieveTopK}(K, q, k)$ // {Top-$k$ demonstrations}
\STATE $\mathbf{M}_0 \leftarrow \mathbf{0}$, $t \leftarrow 1$ // {Initialize memory}
\FOR{each batch $B_j \subset D$}
    \FOR{each $d_i \in B_j$}
        \STATE $\mathbf{M}_t \leftarrow \mathcal{M}(\mathbf{M}_0, d_i)$ 
        \STATE $t \leftarrow t + 1$
    \ENDFOR
    \STATE $\mathbf{M}_t \leftarrow \mathbf{M}_{t-|B_j|} + \frac{1}{|B_j|} \sum_{i=1}^{|B_j|} \mathbf{M}_{t-|B_j|+i}$ // {Update memory}
\ENDFOR
\STATE $output \leftarrow \mathcal{M}(\mathbf{M}_k, q)$ // {Final inference with memory and query}
\RETURN $output$
\end{algorithmic}
\label{alg:alg2}
\end{algorithm}

\section{Experimental Analysis}
\label{sec:appendix}

\subsection{Dataset Statistics} 
\label{appendix:data_statistic}
We evaluate our proposed method on three event argument extraction (EAE) datasets.

\noindent \textbf{RAMS}~\cite{ebner2020multi} is a document-level EAE dataset comprising 9,124 annotated events from English online news. We use a sliding window approach to aggregate events within the same context into single instances with multiple events, following the original train/dev/test split as in~\cite{he2023revisiting}.

\noindent \textbf{WikiEvents}~\cite{zhang2020two} is a document-level EAE dataset with events from English Wikipedia and related news articles. Although it includes co-reference links for arguments, we only utilize the exact argument annotations in our experiments.

\noindent \textbf{ACE05}~\cite{doddington2004automatic} is a labeled corpus for information extraction, including newswire, broadcast news, and telephone conversations. We use the English event annotations for sentence-level EAE, following the preprocessing method described by~\cite{ma2022prompt}.

The detailed dataset statistics for the three datasets are presented in Table~\ref{tab:dataset_characteristics}.

\begin{table}[ht]
\centering
\setlength{\tabcolsep}{1.0mm}{
\begin{tabular}{@{}lccc@{}}
\hline
\textbf{Dataset}      & \textbf{RAMS} & \textbf{WikiEvents} & \textbf{ACE2005} \\ \hline
\textbf{\# Event Types}        & 139  & 50         & 33   \\
\textbf{\# Events per Doc}   & 1.25 & 1.78       & 1.19 \\
\textbf{\# Args per Event}  & 2.33 & 1.40       & 1.35 \\
\midrule
\textbf{\# Total Events}              &      &            &      \\
Training Set               & 7329 & 3241       & 4202 \\
Validation Set                  & 924  & 345        & 450    \\
Test Set                 & 871  & 365        & 403 \\
\bottomrule
\end{tabular}
}

\caption{Overview of Dataset Statistics.}
\label{tab:dataset_characteristics}
\end{table}
\subsection{Implement Details for models in Encoder-Decoder Architecture} 
Our models, including PAIE-R, BART-Gen-R, PAIE-CMR and BART-Gen-CMR, based on encoder-decoder architectures, are run on a single RTX 4090 GPU. All experimental results are averaged over five random seeds. The trainable gating scalar $\gamma$ is initialized to 0 for all layers. The detailed hyperparameters for PAIE-CMR and BART-Gen-CMR are presented in Table~\ref{table:hyper} and Table~\ref{table:hyper2}.

\begin{table}[ht]
\centering
\setlength{\tabcolsep}{0.3mm}{
\begin{tabular}{lccc}
\hline
\textbf{Hyperparameters}  & \textbf{RAMS} & \textbf{Wiki} & \textbf{ACE2005} \\ \hline
Training Steps*                                         & 20000         & 20000               & 15000          \\
Warmup Ratio                                              & 0.1           & 0.1                & 0.2           \\
Learning Rate                 & 2e-5          & 2e-5               & 2e-5          \\
Gradient Accum Steps*                & 8             & 8                  & 8             \\
$\textit{Max}\_\textit{retrieval}$*               & 8             & 8                  & 8             \\
Batch Size                         & 4             & 4                  & 16             \\
Context Window Size              & 250           & 250                & 250           \\
Max Span Length                  & 10            & 10                 & 10            \\
Max Encoder Seq Length           & 500           & 500                & 500           \\
Max Prompt Length              & 210          & 210                & 80    \\
Demonstration Batch Size*              & 4          & 4                & 4
\\ \hline
\end{tabular}}
\caption{Hyperparameter settings for PAIE-CMR. * means that we tuned the hyperparameters in our experiments. The rest of hyperparameters are set the same as PAIE~\cite{ma2022prompt}. }
\label{table:hyper}
\end{table}

\begin{table}[ht]
\centering
\setlength{\tabcolsep}{0.3mm}{
\begin{tabular}{lccc}
\hline
\textbf{Hyperparameters}  & \textbf{RAMS} & \textbf{Wiki} & \textbf{ACE2005} \\ \hline
Training Epochs*                                          & 8         & 8               & 5          \\
Warmup Ratio                                              & 0.0           & 0.0                & 0.0           \\
Learning Rate                 & 3e-5          & 3e-5               & 3e-5          \\
Gradient Accum Steps*                & 8             & 8                  & 8             \\
$\textit{Max}\_\textit{retrieval}$*               & 8             & 8                  & 8             \\
Batch Size                         & 2             & 2                  & 8             \\

Weight Decay           & 0           & 0                & 0     
\\
Demonstration Batch Size*              & 4          & 4                & 4
\\ \hline
\end{tabular}}
\caption{Hyperparameter settings for BART-Gen-CMR. * means that we tuned the hyperparameters in our experiments. The rest of hyperparameters are set the same as PAIE~\cite{huang2023simple}. }
\label{table:hyper2}
\end{table}

\subsection{Implement Details for models in Decoder-Only Architecture} 
\label{sec：appendix_decoder}
\begin{figure*}[htbp]
    \centering
    \includegraphics[width=1.0\linewidth]{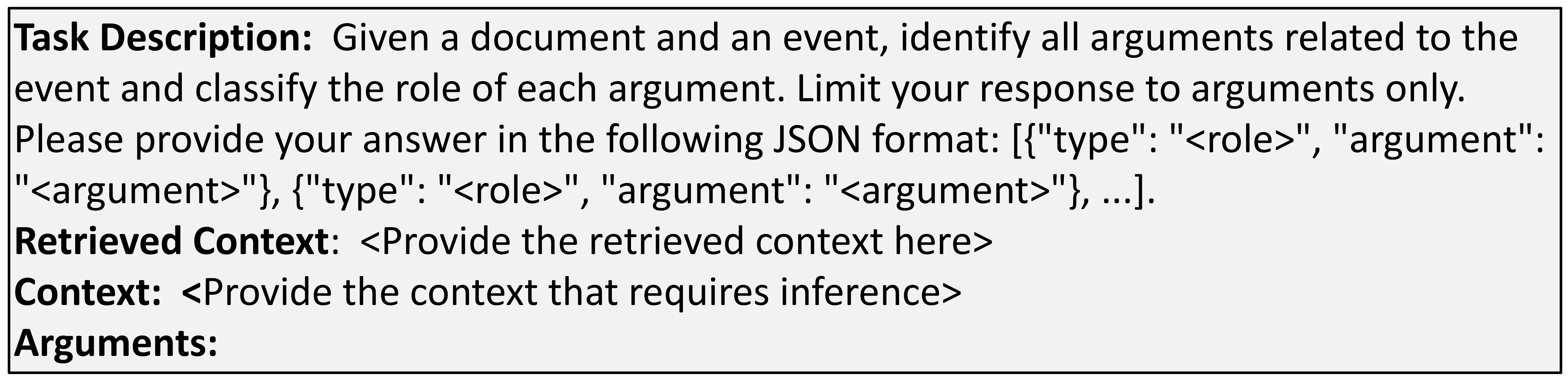}
    
    \caption{Our designed prompt for EAE task for normal decoder-only LLMs.}
    \label{fig:prompt}
\end{figure*}

\begin{figure*}[htbp]
    \centering
    \includegraphics[width=1.0\linewidth]{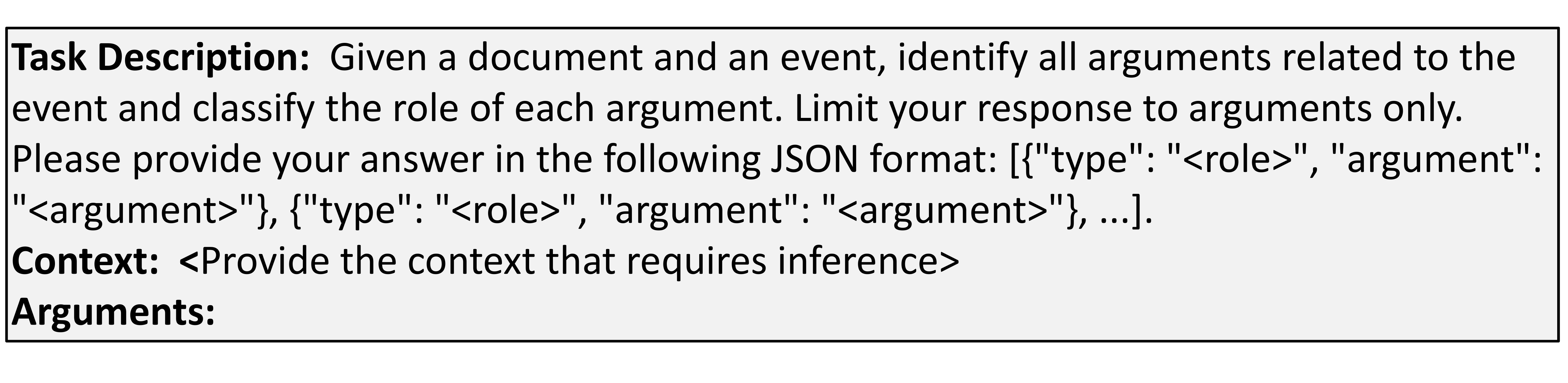}
    
    \caption{Our designed prompt for EAE task for our CMR-based LLMs.}
    \label{fig:prompt_cmr}
\end{figure*}

We choose LLaMA3-8b-instruct for full-parameter fine-tuning on the RAMS dataset. The experiments are conducted using four 80GB A100 GPUs, with training lasting approximately one hour for 3 epochs. The batch size is set to 2 per GPU, with 8 gradient accumulation steps, and the maximum input length is 4096 tokens. 
During the training process, we format the inputs as $\texttt{<bos> X Y <eos>}$ and the labels as $\texttt{<ignore> \ldots <ignore> Y <eos>}$. In this setup, \texttt{<bos>} marks the beginning of the sequence, \texttt{X Y} represents the input context and label, and \texttt{<eos>} indicates the end of the sequence. The labels are structured to ignore the initial part of the sequence (denoted by \texttt{<ignore>} tokens), focusing only on \texttt{Y <eos>} for loss calculation during training.
The prompts are specifically designed for the EAE task, as detailed in Figure~\ref{fig:prompt} and Figure~\ref{fig:prompt_cmr}.  We train the LLaMA3-SFT-CMR model using the CMR mechanism, following the training strategy in Section~\ref{sec:training}. The memory is updated only after the model processes an entire instance.  For comparison, we also train a LLaMA3-SFT model using standard supervised fine-tuning. 

\subsection{Domain Transfer Experiments}
\label{appendix:domain_transfer}
In this section, to simulate a real-world scenario, we explore the capabilities of our model with RAG applied to test sets of new ontologies (event types and argument types),  following the studies by~\cite{lietal2021document, du2022retrieval}. Specifically, we conduct experiments on the RAMS, WikiEvents, and Ace05 datasets, training the model on the source dataset (\texttt{src}) and evaluating it on the target dataset (\texttt{tgt}).
As shown in Table~\ref{exp:domain_trans}, compared to PAIE, our PAIE-CMR performs better in all domain transfer scenarios, demonstrating our model's capability with RAG under new ontologies. This illustrates the robust generalizability of our approach.
\begin{table*}[ht]
\centering
\small{
\setlength{\tabcolsep}{3mm}{
\begin{tabular}{cccccccc}
\hline
\multirow{3}{*}{Model} & RAMS                  & RAMS                 & WIKI                 & WIKI                 & ACE05                & ACE05                & \multirow{3}{*}{Avg} \\
                       & \multicolumn{1}{c}{$\Downarrow$ } & \multicolumn{1}{c}{$\Downarrow$ } & \multicolumn{1}{c}{$\Downarrow$ } & \multicolumn{1}{c}{$\Downarrow$ } & \multicolumn{1}{c}{$\Downarrow$ } & \multicolumn{1}{c}{$\Downarrow$ } &                      \\
                       & WIKI                  & ACE05                & RAMS                 & ACE05                & RAMS                 & WIKI                 &                      \\ \hline
PAIE                   & 20.5                  & 32.4                 & 32.2                 & 48.5                 & 20.3                 & 40.6                 & 32.4                \\
PAIE-CMR (Ours)              & \textbf{26.8}         & \textbf{35.1}        & \textbf{34.9}        & \textbf{51.1}        & \textbf{23.8}        & \textbf{45.8}        & \textbf{36.3}       \\ \hline
\end{tabular}
}
\caption{Performance metrics (Arg-C F1 score) across various \texttt{src}$\Rightarrow$\texttt{tgt} configurations are detailed. The model is trained on the \texttt{src} dataset and evaluated on the \texttt{tgt} dataset. The Avg column reflects the mean scores from all \texttt{src}$\Rightarrow$\texttt{tgt} scenarios.}
\label{exp:domain_trans}}
\end{table*}
\section{Detailed Analysis of Compressive Memory-based Retrieval}
\label{appendix:C}
In this section, we further analyze our CMR mechanism and show that it enables the retrieval of information from demonstrations stored in memory.
First, we briefly introduce the concept of traditional attention and linear attention~\cite{katharopoulos2020transformers} to lay the groundwork for our approach, then we demonstrate that our approach can be considered a natural extension of linear attention and therefore can be seen as a retrieval and extraction of existing information.

 For an embedded input sequence $(\mathbf{x}_1,\mathbf{x}_2,\cdots,\mathbf{x}_N)$, traditional attention machenism generates a sequence-to-sequence mapping by calculating the interactions between inputs from each location and inputs from other locations and integrating them into its own representation, obtaining the output sequences $(\mathbf{y}_1, \mathbf{y}_2,\cdots, \mathbf{y}_N)$,  
 Taking the $i$-th token as an example,and disregarding the scaling factor,the resulting output $\mathbf{y}_i$ of the aggregated global information is as follows :

 \begin{equation*}
    \mathbf{y}_{i}=\frac{\sum_{j=1}^Nexp(\mathbf{q}_i\mathbf{k}_j^T)\mathbf{v}_j}{\sum_{j=1}^Nexp(\mathbf{q}_i\mathbf{k}_j^T)}.
\end{equation*}
Here,$\mathbf{q}_i, \mathbf{k}_i, \mathbf{v}_i \in \mathbb{R}^{1\times d}$, correspond to the $i$-th token's  query, key, and value in traditional attention. The softmax function $\frac{exp(\mathbf{q}_i\mathbf{k}_j^T)}{\sum_{j=1}^Nexp(\mathbf{q}_i\mathbf{k}_j^T)}$ can be viewed as a weighting coefficient based on the similarity between $x_i$ and $x_j$. ~\cite{katharopoulos2020transformers} treat this similarity calculation method as one of the general functions $sim(\cdot,\dots)$ representing the interactions between different tokens. Linear attention uses a kernel function $\mathcal{K}$ to represent the $sim(\cdot,\cdot)$, i.e $sim(\mathbf{q}_i,\mathbf{k}_j):=\mathcal{K}(q_i,k_j)=\sigma(\mathbf{q}_i)\sigma(\mathbf{k}_j^T)$, here $\sigma:\mathbb{R}^{1\times d} \to \mathbb{R}^{1\times d'}$ is a nonlinear and positive map~\cite{tiezzi2024state,tsai2019transformer}. Then the output can be written by the following formula:
\begin{equation*}
    \mathbf{y}_{i}=\frac{\sum_{j=1}^Nsim(\mathbf{q}_i,\mathbf{k}_j)\mathbf{v}_j}{\sum_{j=1}^Nsim(\mathbf{q}_i,\mathbf{k}_j)}=\frac{\sum_{j=1}^N\sigma(\mathbf{q}_i)\sigma(\mathbf{k}_j^T)\mathbf{v}_j}{\sum_{j=1}^N\sigma(\mathbf{q}_i)\sigma(\mathbf{k}_j^T)}.
\end{equation*}

The function $\sigma$ in linear attention serves to replace the traditional attention mechanism based on softmax's similarity. The splitting of the $sim(\cdot,\cdot)$ allows the calculation order of Q, K, and V to be swapped, so that the complexity of the calculation does not need to increase with the quadratic complexity of the sequence length. For details, please refer to ~\cite{katharopoulos2020transformers}.

\begin{table}[]
\centering
\setlength{\tabcolsep}{1mm}{
\begin{tabular}{ccccl}
\hline
\multirow{2}{*}{\textbf{Method}} & \multirow{2}{*}{\textbf{\#N }} & \multirow{2}{*}{\textbf{\#Demo BS}} & \multicolumn{2}{c}{\textbf{RAMS}}      \\ \cline{4-5} 
                                 &                               &                                        & \multicolumn{2}{c}{Inference Time (s)} \\ \hline
PAIE                             & 0                             & -                                      & \multicolumn{2}{c}{22.95}              \\ \hline
\multirow{7}{*}{PAIE-CMR}        & 1                             & 1                                      & \multicolumn{2}{c}{46.18}              \\
                                 & 5                             & 1                                      & \multicolumn{2}{c}{136.21}             \\
                                 & 5                             & 4                                      & \multicolumn{2}{c}{90.72}            \\
                                 & 10                            & 1                                      & \multicolumn{2}{c}{227.26}             \\
                                 & 10                            & 4                                      & \multicolumn{2}{c}{141.75}             \\
                                 & 15                            & 1                                      & \multicolumn{2}{c}{356.44}             \\
                                 & 15                            & 4                                      & \multicolumn{2}{c}{206.32}             \\ \hline
\end{tabular}
}
\caption{Inference time (second) for PAIE and PAIE-CMR on the
test set of RAMS dataset. Experiments
are run on one same RTX 4090 GPU. \# Demo BS denotes the batch size of processing demonstrations.}
\label{tab:inference}
\end{table}

Our work generalizes this computation method from vectors to matrices and realizes information aggregation from tokens to the whole text. 
Combined with the equations \ref{eq:mem_update1} and \ref{equation:retrieval}, $\mathbf{A}_\text{ret}$ can be represented by the following formula:
\begin{align*}
\mathbf{A}_{\text{ret}} = \frac{\sigma(\mathbf{Q}) \mathbf{M}_{k}}{\sigma(\mathbf{Q}) \mathbf{n}_{k}}=\frac{\sigma(\mathbf{Q}) \sum_{i=1}^k\sigma(\mathbf{K}^{d_i})^T\mathbf{V}^{d_i}}{\sigma(\mathbf{Q}) \mathbf{n}_{k}}  \\
=\frac{\sum_{i=1}^k\sigma(\mathbf{Q}) \sigma(\mathbf{K}^{d_i})^T\mathbf{V}^{d_i}}{\sigma(\mathbf{Q}) \mathbf{n}_{k}}.
\label{equation:}
\end{align*}
Here, it can be considered that \(\sigma(\mathbf{Q})\sigma(\mathbf{K}^{d_i})^T\) is the approximation of the \(sim(\cdot,\cdot)\) function acting on the matrix, representing the "similarity" between the query \(\mathbf{Q}\) and each demonstration \(d_i\). Understanding operations between matrices that result in a new matrix \(\sigma(\mathbf{Q})\sigma(\mathbf{K}^{d_i})^T\) rather than a single value using the concept of "similarity" may be unreasonable. Known that our approach involves giving each existing demonstration \(d_i\) interaction with the query \(\mathbf{Q}\), closely related to the demonstration \(d_i\) itself, this whole process can be understood through a selection mechanism~\cite{gu2023mamba}: retaining important information among  $\{d_1, d_2, \dots, d_k\}$ related to the query \(\mathbf{Q}\) and discarding unimportant information. A function $f(Q,d_i)=\sigma(\mathbf{Q})\sigma(\mathbf{K}^{d_i})^T$ determines the importance of the demonstration $d_i$, influences how the representation $\mathbf{V}^{d_i}$ acts on the final representation of input, i.e.$A_{ret}$. Therefore, this process can be viewed as the query \(\mathbf{Q}\) retrieving information from the candidate demonstrations. Our algorithm bears some resemblance to linear attention~\cite{katharopoulos2020transformers}, the following outlines the key difference between these two models: while linear attention seeks to map each token's feature vector to another vector that consolidates all tokens' information, our model aims to aggregate existing text information (matrix but not vector) using an operation method similar to linear attention, the aggregated information (information of demonstrations) is then integrated into new input text to derive a new feature representation. 

\section{Efficiency Analysis}

In this section, we explore the efficiency of the CMR mechanism. We compare the inference time of PAIE-CMR and PAIE on the RAMS test set. For PAIE-CMR, we measure the time required to retrieve 1, 5, 10, and 15 demonstrations. The inference batch size is set to 1, and the demonstration batch size \(B_j\) is 4.

As shown in Table~\ref{tab:inference}, our PAIE-CMR model increases inference time compared to PAIE due to the need to store demonstrations. However, this additional time is justified by the corresponding improvement in performance. Moreover, by processing demonstrations in batches, our approach effectively reduces the overall time cost during inference.

\begin{figure*}[htbp]
    \centering
    \includegraphics[width=1.0\linewidth]{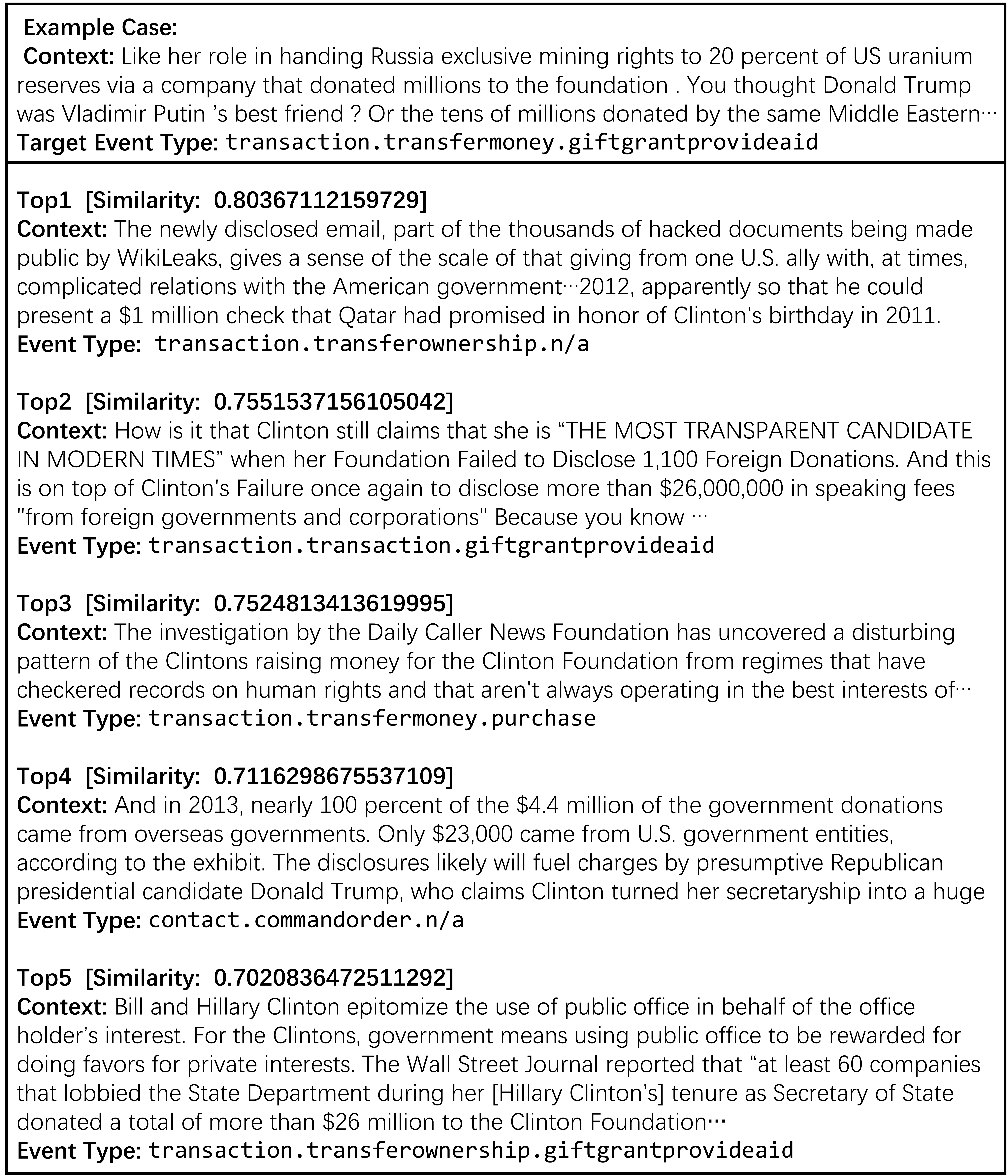}
    
    \caption{A specific case from the RAMS dataset highlighting the importance of diversity in demonstrations.}
    \label{fig:diversity demo}
\end{figure*}  

\section{Demonstration Diversity Analysis}
In this section, we analyze the improvement in diversity when retrieving multiple demonstrations compared to retrieving only the top 1 demonstration. We provide a specific case to illustrate this. As shown in Figure~\ref{fig:diversity demo},  the example case is an instance randomly selected from the RAMS dataset. Below are the demonstrations retrieved using SBERT based on similarity. It is evident that retrieving the top 5 demonstrations, compared to just the top 1, results in a greater diversity of event types. A more diverse set of demonstrations can provide richer retrieval information, ensuring the effectiveness of RAG.

\end{document}